\documentclass[letterpaper, 10 pt, journal, twoside]{ieeetran}
\pdfoutput=1 
\IEEEoverridecommandlockouts                              %

\usepackage{graphics} %
\usepackage{epsfig} %
\usepackage{mathptmx} %
\usepackage{times} %
\usepackage{amsmath} %
\usepackage{amssymb}  %
\usepackage{url}
\usepackage{diagbox} 
\usepackage{float}%

\usepackage{graphicx}
\usepackage{bm}
\usepackage{xcolor, soul}
\usepackage{float}
\usepackage{subfigure}
\usepackage{multirow}
\usepackage{makecell}
\usepackage[ruled,linesnumbered]{algorithm2e}
\usepackage{algpseudocode}
\usepackage{scalerel}
\usepackage{tikz}
\usepackage[labelfont=bf]{caption}
\usepackage{pifont}
\usepackage{booktabs}
\usepackage{multirow}

\makeatletter
\let\NAT@parse\undefined
\makeatother
\usepackage[colorlinks=true,linkcolor=blue,citecolor=blue]{hyperref} %

\hypersetup{
  colorlinks   = true,    %
  urlcolor     = blue,    %
  linkcolor    = blue,    %
  citecolor    = blue      %
}

\title{\LARGE \bf
Encountered-Type Haptic Display \textit{via} Tracking Calibrated Robot
}
\author{Chenxi Xiao, Yuan Tian
\thanks{ Chenxi Xiao is with the Industrial Engineering at Purdue University, 
        {\tt\footnotesize xiao237@purdue.edu}}%
\thanks{ Yuan Tian is with the Computer Science at Purdue University, 
        {\tt\footnotesize tian211@purdue.edu}}%
}

\begin{document}

\maketitle

\begin{abstract}
In the past decades, a variety of haptic devices have been developed to facilitate high-fidelity human-computer interaction (HCI) in virtual reality (VR). In particular, passive haptic feedback can create a compelling sensation based on real objects spatially overlapping with their virtual counterparts. 
However, these approaches require pre-deployment efforts, hindering their democratizing use in practice.
We propose \textit{Tracking Calibrated Robot} (TCR), a novel and general haptic approach to free developers from deployment efforts, which can be potentially deployed in any scenario. 
Specifically, we augment the VR with a collaborative robot that renders haptic contact in the real world while the user touches a virtual object in the virtual world. 
The distance between the user's finger and the robot end-effector is controlled over time. The distance starts to smoothly reduce to zero when the user intends to touch the virtual object.
A mock user study tested users' perception of three virtual objects, and the result shows that TCR is effective in terms of conveying discriminative shape information.
\end{abstract}

\section{Introduction}
In virtual reality (VR), haptic feedback enables users to interact with the virtual world with augmented fidelity. 
Haptic feedback is used to provide tactile sensations and enhance the immersive experience of VR. By simulating touch, haptic feedback makes VR interactions feel more real and tangible to the user, supplementing visual and audible sensations.
Aiming to create such feedback, it has motivated the development of a variety of haptic rendering technologies including active haptics controllers \cite{blake2009haptic}, passive feedback and retargeting techniques \cite{cheng2017sparse}.

To generate haptic feedback, passive approaches have shown the effectiveness of providing compelling sensations. The haptic feedback is generated by making contact with a real scene, which has a spatial correlation with its virtual counterparts. However, such technology requires the corresponding real objects to be deployed ahead of time, which lacks the flexibility to be generalized to general scenes with complex object curvatures, and dynamic objects. For instance, changes made to virtual objects can require time-consuming changes to their physical passive-haptic counterparts \cite{kohli2010redirected}. While this limitation can be partially relieved by wrapping and retargeting, the inconsistency could still be perceivable if the difference is significant. 

To extend the application of passive haptic feedback to general scenes, the proposed work aims to propose and implement a novel approach. To be specific, the haptic feedback is provided by physical contact with objects in the real world. However, the main difference is that the contact sensation is conveyed through a robot dynamically, which moves a prop towards the user's fingertip when emulating a contact. Compared to the existing passive approaches, we hypothesize that the proposed approach has the following advantages. First, the robot system could provide the flexibility for emulating versatile scenes. As a result, the deployment of stationary props may not be needed. Second, the proposed approach can simulate the interaction processing with dynamic objects, such as frictional and damping effects when pushing an object. Third, by reorientating a real prop (such as a board), different contact forces could be generated on the user's fingertip, which allows for the simulation of different surface curvatures. Last, the proposed approach can simulate large surfaces when transiting the end-effector with the user's hand motion.



To implement the above idea, the proposed work is divided into two submodules. The first module aims to control a collaborative robot that is used to convey skin contact. The second module is a virtual reality program that tracks the relative spatial relationship between the user's hand movement and an object to be interacted with. This will be explained in detail in the methodology section.

\section{Related Work}
\subsection{Passive Haptic Feedback}
Passive haptic feedback is a feedback mechanism that utilizes physical objects to simulate interactions in a virtual environment. To be specific, this type of feedback is generated through a mapping proxy between virtual objects and tangible objects in the real world. For instance, haptic effects can be replicated when touching virtual objects on a desk \cite{cheng2017sparse}. Similarly, a handheld golf club prop can recreate the haptic experience in a virtual golf game \cite{franzluebbers2018performance}.

The advantage of passive feedback lies in its ability to provide high-fidelity sensations derived from real props, which often feature intricate details. However, despite its effectiveness, obtaining the necessary real-world objects to replicate haptic cues may not always be straightforward. As these real-world objects must be arranged in advance by humans, it requires human effort to locate and create them. This is especially true when the simulated objects lack real-world counterparts (potentially require additional manufacturing work) or with large dimensions (making deployment challenging). Therefore, the utilization of passive haptic feedback remains restricted \cite{cheng2017sparse}.

To enhance the versatility of passive haptic feedback, active haptic feedback has been adopted. This form of active haptic feedback is developed using devices such as gloves \cite{blake2009haptic} and force feedback controllers \cite{massie1994phantom}. However, such devices are usually based on specialized actuators. Given that these actuators are typically worn by the operator, their usage can be cumbersome and potentially limit mobility. As a solution, there is a need for an active haptic rendering approach in which the devices are loosely coupled to human operators, such as Encountered-type haptic displays.

\subsection{Encountered-type haptic disdplays}
Encountered-type haptic displays (ETHDs) has been developed to create haptic feedback, which provide haptic sensation by actively positioning encountered-type objects at proper times and locations \cite{mercado2021haptics}. Compared to worn haptic devices (such as gloves) or passive haptic displays, this technology facilitates a natural way of interacting with objects while avoiding the complexity of scene deployment. The hardware scheme of the ETHD system is composed of an actuated haptics display module and the corresponding Virtual Reality device. When the user interacts with a virtual object, the actuator delivers the haptic display module to the corresponding location in the real world, which allows realistic touch events. This location usually changes according to the user's intention, which is predictable based on the user's hand or body tracking data. 

Over the past decades, a wide variety of ETHDs systems has been developed. Based on the application, the system design varies according to the type of surface display, actuator, and the type of body/hand tracking devices. For instance, The locomotion can be generated by a collaborative robot, or a drone \cite{fedoseev2020teslamirror, kim2020synthesizing, mercado2019entropia, abdullah2017hapticdrone, yamaguchi2016non}. Different props can be used to deliver different sensations of interaction. A board can be used to render the contact force with a wall \cite{kim2020synthesizing}. Real props that are rich in features can render complex features such as object edges \cite{mercado2019entropia}.



\section{Method} \label{sec:method}
In this section, we describe the approach and implementation of the tracking calibrated robot (TCR). The key idea is to generate contact with a user's hand in the real world using a collaborative robot when a virtual contact happens in the virtual world correspondingly. Here we leverage a robot as the haptic device, which can change the position of its end effector to interact with the user's hand, or to change the pose of the end effector to generate a variety of contact forces. This allows us to simulate a wide range of haptic applications including point contact, sliding motion, or force-based object interaction. 

\subsection{System architecture}
\begin{figure}[htb]
    \centering
    \includegraphics[width=\linewidth]{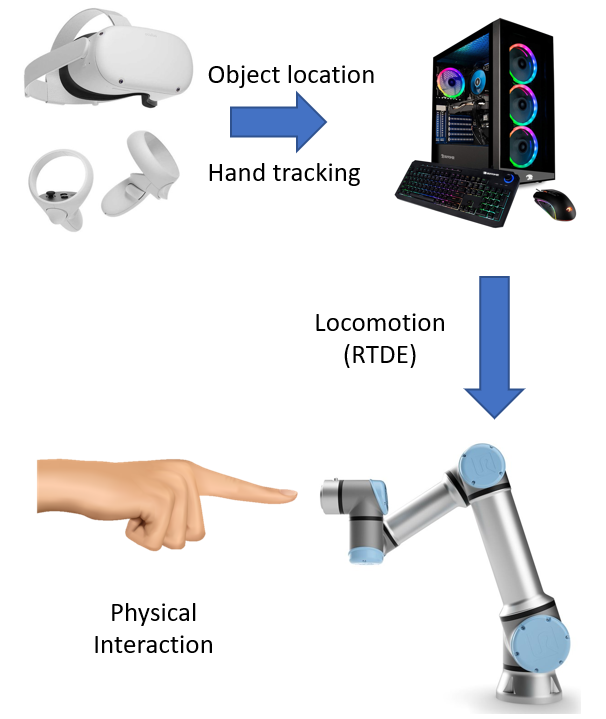}
    \caption{The diagram of the system architecture. }
    \label{fig:sys_diagram}
\end{figure}

The system consists of the following components. First, a collaborative robot is used to provide the interaction events in the real world. Here we leverage the UR16e robot, a collaborative robot with a reach radius of 0.9m. Using a collaborative robot benefits in terms of safety because its moving velocity as well as the interaction force could be monitored and controlled in real time. Besides, having the force information enables it to provide haptic feedback for simulating object interaction, such as pushing an object or simulating the damping effect of a mass-spring system. The robot could be remotely controlled by Real-Time Data Exchange (RTDE) protocol over a standard TCP/IP connection. Second, Oculus Quest, a VR HMD device is used to provide a virtual experience to the user. For simplicity, we assume the user's fingertip location could be substituted by the location of the controller plus a small offset. Besides, a desktop PC with an Ubuntu 18.04 system bridges the headset and the robot, providing flexibility for state control, data logging, and real-time processing. The system architecture is shown in Fig.~\ref{fig:sys_diagram}.

\subsection{User interaction with the robot}
In this section, we introduce how a robot generates haptic feedback by interacting with the user's hand. While interacting with the whole hand is desired, the current robot system does not have a sufficient degree of freedom to simulate a pressure distribution on the skin surface. Therefore, the process for generating such interaction is studied on a point only, which is at the user's fingertip in our case. 

In the virtual world, we define the location of the user's fingertip as a point $\mathbf{x}^v_f \in \mathbb{R}^3$, and a proximity point on the object surface as $\mathbf{x}^v_o  \in \mathbb{R}^3$. The Euclidean distance between two points is expressed as $||\mathbf{x}^v_f - \mathbf{x}^v_o||$. Correspondingly, in the real world, the user's fingertip is located at $\mathbf{x}^r_f \in \mathbb{R}^3$, and the robot end effector is located at $\mathbf{x}^r_o \in \mathbb{R}^3$. A sufficient condition for simulating the virtual contact is to find $\mathbf{x}^r_o$ that could satisfy: $||\mathbf{x}^v_f - \mathbf{x}^v_o|| - ||\mathbf{x}^r_f - \mathbf{x}^r_o|| = 0$. This is because a real contact will always be generated when $||\mathbf{x}^v_f - \mathbf{x}^v_o||=0$ in the virtual world. Note that there is an infinite number of $\mathbf{x}^r_o$ that could satisfy this constraint. Therefore, the selection of $\mathbf{x}^r_o$ does not necessarily be aligned with $\mathbf{x}^v_o$ exactly. 

Consider the reachability of the robot, the $\mathbf{x}^r_o$ is chosen to be in front of the user, i.e., satisfies $(\mathbf{x}^r_o-\mathbf{x}^r_f) \cdot (\mathbf{O}^r_{robot}-\mathbf{O}^r_{headset})>0$, where $\mathbf{O}^r_{robot}$ and $\mathbf{O}^r_{headset}$ are the origin points of robot and headset in the world axis, respectively. By letting the robot end-effector be always in front of the user, it avoids collision with the user's arm. For simplicity, the degree of freedom for controlling $\mathbf{x}^r_o$ is assigned to be along the vector $\mathbf{O}^r_{robot}-\mathbf{O}^r_{headset}$. The other components are subject to the user's hand location i.e., always be in front of the user's fingertip. 

A calibration step is required to align the virtual world with the robot coordinate axis. This is subject to the definition of orthogonal Procrustes problem, which is given as Eq.~(\ref{eq:procrusts}):
\begin{equation}
\mathbf{R}=\arg \min \|\Omega \mathbf{A}-\mathbf{B}\|_{F} \quad \text { subject to } \quad \Omega^{T} \Omega=I
\label{eq:procrusts}
\end{equation}
Where $\mathbf{A}$ and $\mathbf{B}$ are centered point cloud sets sampled from the VR controller, and the robot end-effector, respectively. The solution to the orthogonal Procrustes problem could be solved by singular value decomposition (SVD) of matrix $\mathbf{B}^T \mathbf{A}$, which is defined as Eq.~(\ref{eq:svd}):

\begin{equation}
\begin{gathered}
\mathbf{M}=\mathbf{B} \mathbf{A}^{T} \\
\mathbf{M}\rightarrow \mathbf{U} \Sigma \mathbf{V}^{T} \\
\mathbf{R}=\mathbf{U} \mathbf{V}^{T}
\end{gathered}
\label{eq:svd}
\end{equation}

Once the optimal rotation $\mathbf{R}$ is obtained, the transition could be trivially obtained as $\mathbf{t}=\mathbf{\hat{B}}-\mathbf{R} \mathbf{\hat{A}}^T$, where $\hat{\mathbf{A}}$ and $\hat{\mathbf{B}}$ are raw (i.e., not centered) point cloud sampled from VR controller, and the robot end-effector, respectively.

\subsection{Design of virtual reality environment}

A virtual environment is created in Unity to convey the visual effects to the user. The scene is composed of 1) a user, and 2) a virtual object to be interacted with. Since the IP address of the server may change over time, the user could modify the server's IP address inside the virtual environment through the text field. The connection is established when the user clicks on a button.

\begin{figure}[htb]
    \centering
    \includegraphics[width=0.8\linewidth]{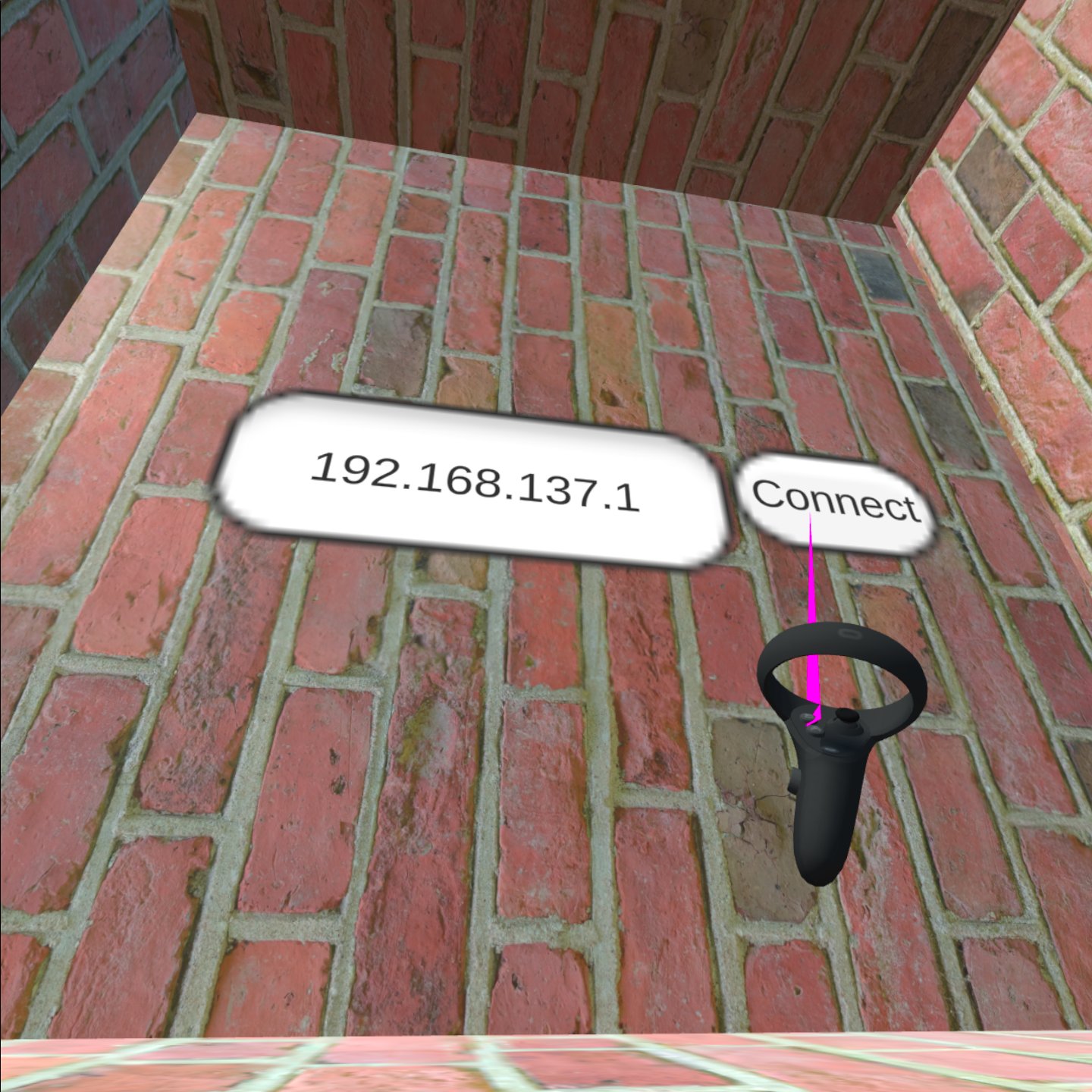}
    \caption{Connect to the server in VE.}
    \label{fig:IP_button}
\end{figure}

\begin{figure}[htb]
    \centering
    \subfigure[]{\includegraphics[width=0.48\linewidth]{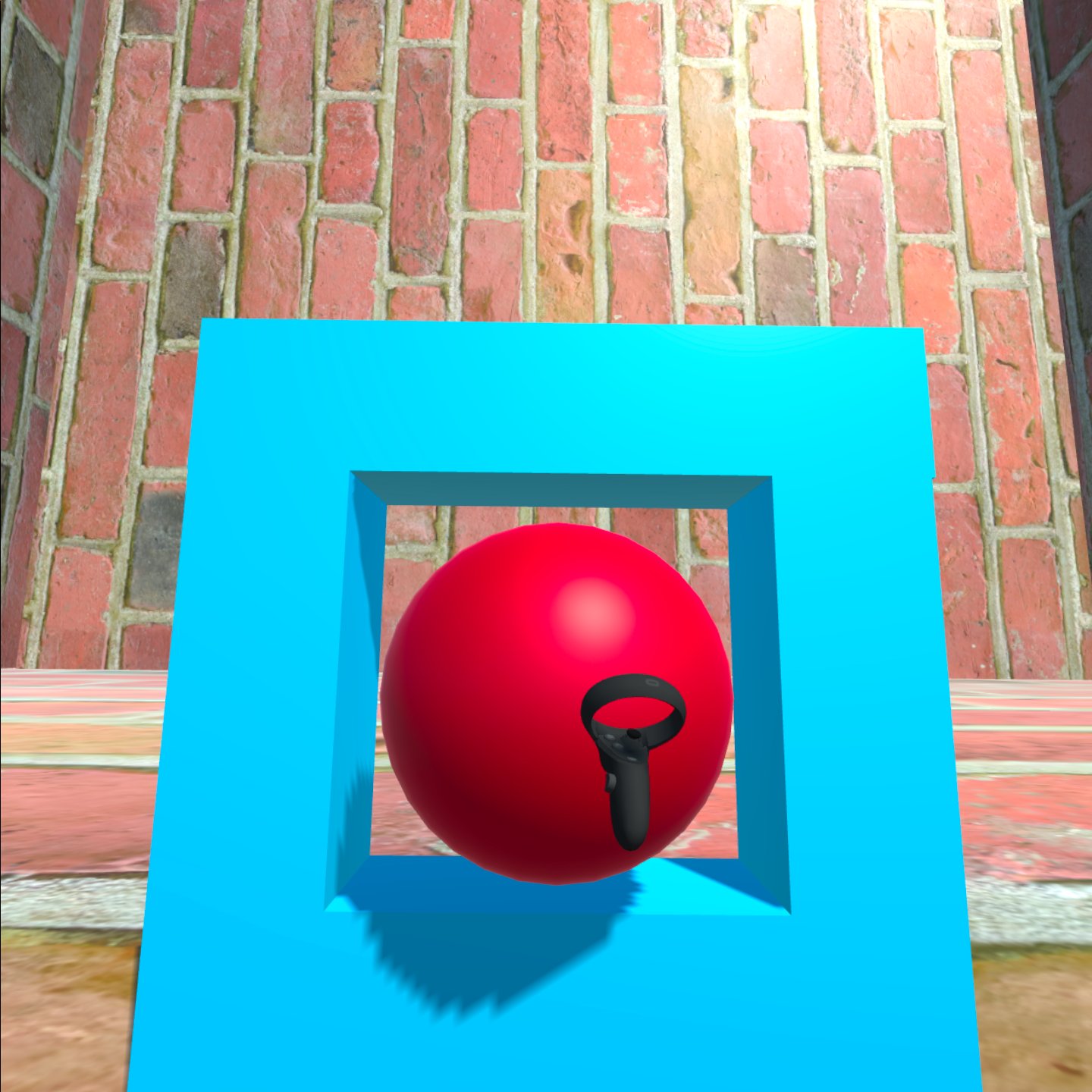}}
    \subfigure[]{\includegraphics[width=0.48\linewidth]{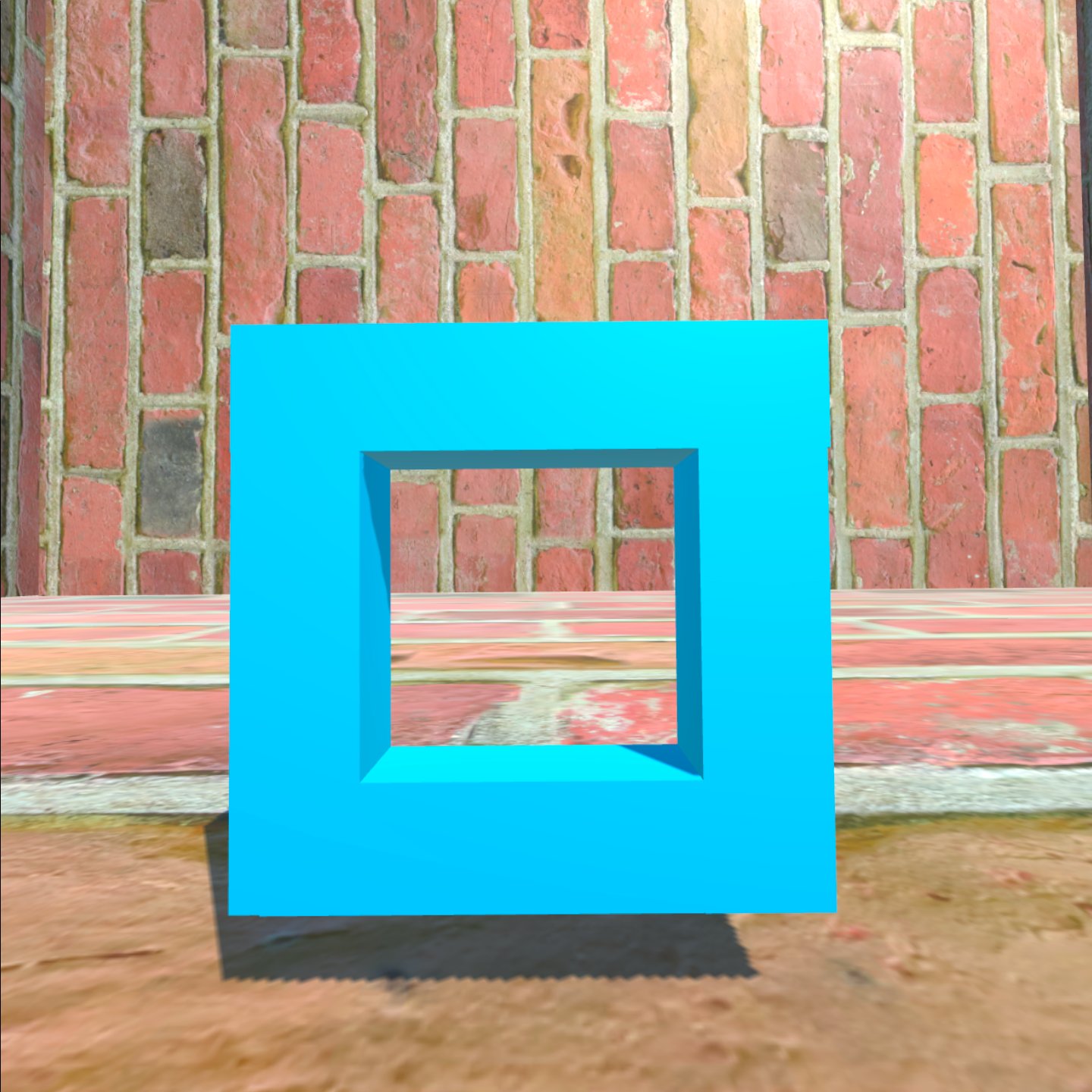}}
    \caption{(a) Interact with the virtual object inside a blue frame. (b) The virtual object is made to be invisible.}
    \label{fig:virtual_interaction}
\end{figure}

After connecting to the server, the user is prompted to interact with the red virtual object inside the blue frame, as shown in Fig.~\ref{fig:virtual_interaction} (a). The user can touch the virtual object using the right hand with the Oculus controller. The client, which is the Oculus headset, will keep tracking the user's right controller as well as calculating the distance between it and the nearest front projected surface, and continuously send the distance as well as the controller's location to the server through a JSON format message. Based on the real-time data, the robot is able to track the user's hand and generate the corresponding contact events.  

In the virtual environment, we currently have included 3 potential objects that the user could interact with. They are: 1) a vertical plane, 2) a sphere, 3) a 45$^{\circ}$ rotated cube, as shown in Fig.~\ref{fig:task1}. By pressing different controller buttons, the user has the ability to 1) randomly switch the virtual object, and 2) make the virtual object invisible, as shown in Fig.~\ref{fig:virtual_interaction} (b).

\section{Experiment} 
\subsection{Calibration}
\begin{figure}[htb]
    \centering
    \includegraphics[width=\linewidth]{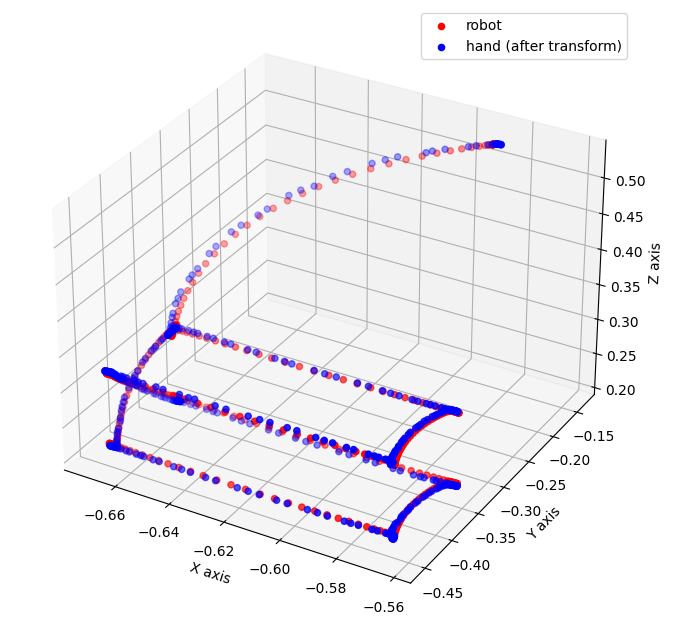}
    \caption{Result of aligning VR coordinate system with the robot coordinate system. Blue dots are the hand positions after being transformed by the calculated homography. Red dots are the positions of the robot end-effector}
    \label{fig:procrustes_result}
\end{figure}

The result of the calibration process is provided. To achieve this, 486 pairs of points were collected. The aligned point cloud is shown in Fig.~\ref{fig:procrustes_result}. It can be observed that the two point clouds have been aligned by the transformation calculated. The alignment error, when being quantified by the Mean Squared Error (MSE), is 0.00485 m. This error value is relatively small compared to the motion range, which is usable in the following applications.

A limitation is that the user needs to recalibrate the system when the VR coordinate system loses track of the real world, which usually happens when the program restarts or when the user removes the headset. This induces increased human efforts in the following user studies.

\subsection{Mock User Study}
Two mock user study was conducted to test the fidelity and usability of our approach.
\subsubsection{Task I}
\begin{figure}[htb]
    \centering
    \subfigure[]{\includegraphics[width=0.3\linewidth]{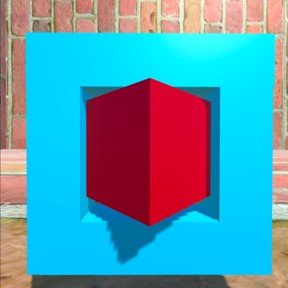}}
    \subfigure[]{\includegraphics[width=0.3\linewidth]{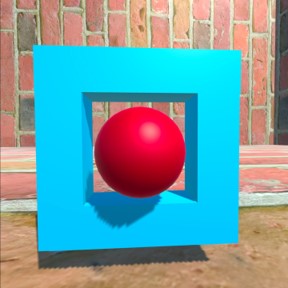}}
    \subfigure[]{\includegraphics[width=0.3\linewidth]{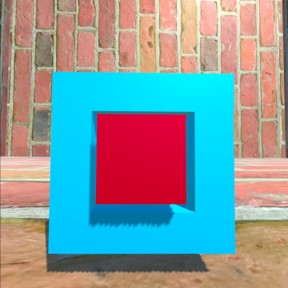}}
    \caption{3 types of virtual objects that are placed in the virtual reality environment. (a) cube, (b) sphere, (c) plane.}
    \label{fig:task1}
\end{figure}

In the first task, we put a virtual object in front of the user. There were three types of virtual objects in total, namely cube, sphere, and plane (refer to Fig.~\ref{fig:task1}). The goal of this task is to recognize the object, in order to prove our assumption that the haptics interface could convey discriminative geometric features about the object. The ground truth object was randomly selected. In the experiment, this object was set to transparent, so that the user would not be able to recognize it through the visual feedback. To guess what object it is, the user needs to infer the shape of this object by touching it.  The haptic feedback is conveyed by the approach through the robot, which is described in Sec.~\ref{sec:method}. After making the guess, the ground truth object is shown to the user.

\begin{figure}[htb]
    \centering
    \includegraphics[width=\linewidth]{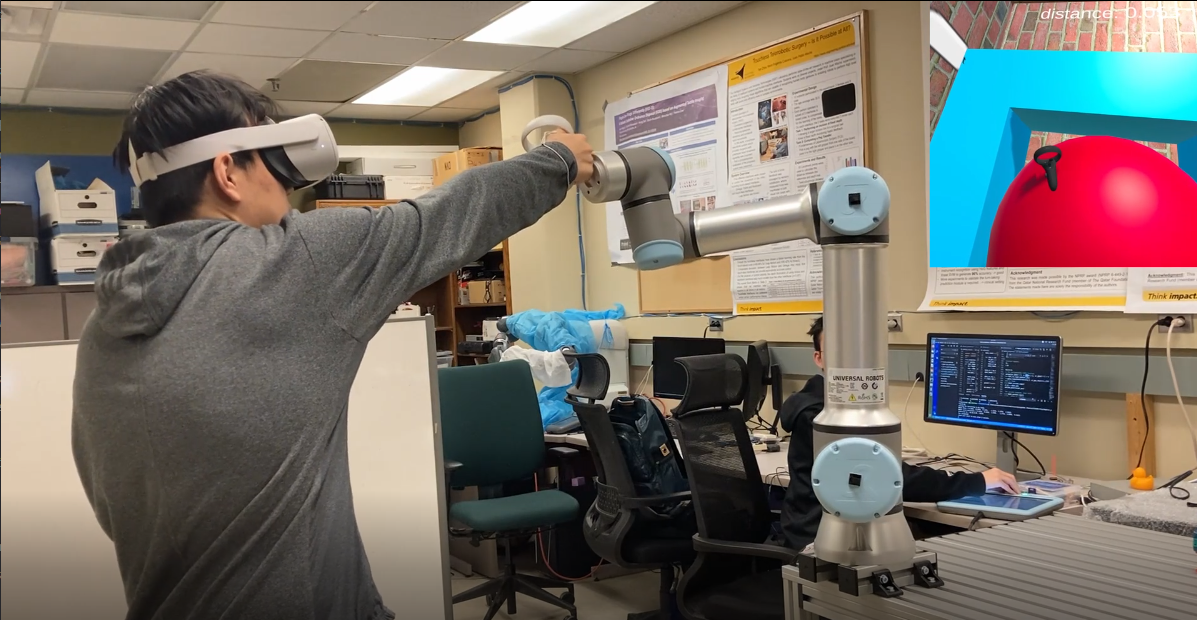}
    \caption{Experiment configuration of the mock user study. The virtual scene is shown in the top right corner.}
    \label{fig:task1_lab}
\end{figure}

\begin{figure}[htb]
    \centering
    \includegraphics[width=0.8\linewidth]{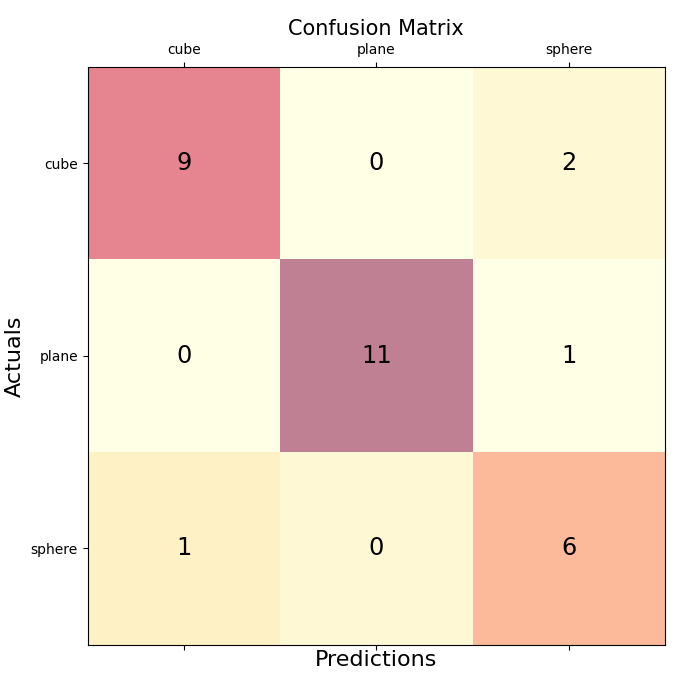}
    \caption{Confusion matrix of the virtual object recognition.}
    \label{fig:task1_confusion}
\end{figure}

The mock user study is conducted with two participants, with 30 evaluation trials in total. An example photo for this experiment is shown in Fig.~\ref{fig:task1_lab}. We have observed 26 success trials, which takes around 86.7\%. This accuracy proves that the system could convey discriminative features about the object. We also visualize the confusion matrix, as shown in Fig.~\ref{fig:task1_confusion}. Here we refer to the controlled condition as the plane case, which has 100\% precision and $\frac{11}{12}$ recall. The experimental conditions are those non-smooth surfaces, i.e., cube, sphere. It can be seen that the complex object shape led to confusion, leading to misclassifications. 
The most common mistake is that the plane, or cube, can be misrecognized as a sphere. We believe this is due to the shape similarity between the cube and the sphere.

\subsubsection{Task II}
In the second task, we aim to test whether the user can slide on the object surface. In this task, the user will try to draw a trajectory on the object's surface. A trajectory segment will be recorded when a button in the controller is pressed, and visualized on the computer screen. 

\begin{figure}[htb]
    \centering
    \includegraphics[width=0.8\linewidth]{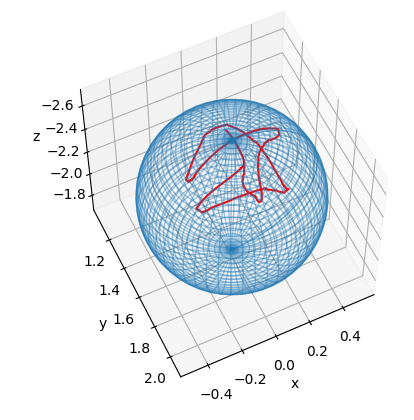}
    \caption{The user hand trajectory (red) overlaid on the surface of the virtual object surface (blue).}
    \label{fig:sphere}
\end{figure}

Without loss of generality, a sphere object in the virtual reality world is used for the experiment. This is shown in Fig.~\ref{fig:sphere}. The average error distance between the user's hand and the sphere is around 2.5 cm.

\section{Limitations \& Future work}

\subsection{Render contacts from arbitrary directions}

For simplicity and safety considerations, we currently only assume the virtual object is in front of the user. However, this is a limitation of the system since the motion range of the robot is limited because it is not possible to render objects behind the user.

As one of the future work, we propose to make the robot contact the user from an arbitrary direction. The main challenge is how to move the robot efficiently while reducing the risk that the robot may collide with the user's body. This problem remains to be challenging since the full-body collision avoidance algorithm for a robotic system in a dynamic environment is still an open problem to be addressed.

\subsection{Render frictional force}
Another issue identified is that when rendering sliding effects onto the object surface, the relative velocity between the robot and the user's hand is not consistent with the friction force direction in the virtual world. This is because the robot needs to track the user's hand using its velocity. It is not clear so far whether it is possible to use the current system to achieve the subgoal of rendering frictional force, while at the same time satisfying the positional constraint.

\subsection{Reducing latency \& User intention prediction}


The current method is based on tracking the location of the Oculus controller, which induces latency between the robot and the hand. The latency of the haptic feedback is determined by the hardware. While this can be reduced by increasing the control gain or increasing the maximum velocity of the robot, latency always exists. Potential issues to this solution are that 1) a high locomotion speed could induce a high impact force when in contact due to inertia, which could be unnatural and reduce the fidelity, 2) potential safety issues.

A potential solution to this problem is to leverage user-intention prediction. For instance, using the user's hand pose, body gesture, and other optical information from the scene, it may be possible to determine the object that the user intends to interact with. This makes it possible to place the robot at the target location ahead of time, which could be an improvement compared to the current solution based on real-time hand tracking. We envision such system could be implemented by predicting a target location through machine learning models, based on processing the user's intention and other profiles collected ahead of time.

\subsection{Haptic feedback through props}
Currently, we are using a 3D-printed plate (PLA material) as the prop. In future work, it is possible to add different kinds of props to the robot. To create a smooth sensation during sliding motion, it is possible to leverage a ball attached to a universal joint.  Another idea is to further enhance the haptic feedback using a vibration motor.

\subsection{More complicated scenario with multiple robots}

With multiple robots working cooperatively, we can fulfill more complicated scenarios. First, it enlarges the working area. In our current design, the workspace is limited by the moving range of the robot. If the virtual object is too large, it may lead to infeasible solutions that are not able to be reached. Second, multiple robots can simulate multiple contacts at the same time. Third, when collision avoidance with the user's body is needed, it reduces the complexity of robot planning by dividing the workspace into a few subregions, with each subregion assigned to a robot. 



\section{Conclusions}

In this report, we present TCR, a novel encountered-type haptic display system that can convey the object's shape. A virtual environment is created through an Oculus HMD headset, while the locomotion system is implemented based on a UR16e robot. We demonstrate that TCR could be used to convey object shape effectively. In our user study, the user can discriminate the object category by making discrete touches on a surface. We also demonstrate that the user could slide on the object's surface, allowing for potential applications such as drawing. 

\section{Acknowledgement}
Chenxi Xiao contributed to the idea, of robotic control, the communication protocol, documents, and data analysis.  
Yuan Tian designed a VR scene for the user study, worked on the communication protocol, and documents, and proposed future work. 
We believe the contribution is equally divided. 

This project was originally developed in Purdue's CS 590 (Introduction to AR/VR) class. We sincerely thank Prof. Voicu Popescu for the valuable knowledge transfer.


{
\small
\bibliographystyle{ieee_fullname}
\bibliography{egbib}
}

\end{document}